\documentclass{article}

\usepackage[preprint]{neurips_2024}

\usepackage[utf8]{inputenc} 
\usepackage[T1]{fontenc}    
\usepackage{hyperref}       
\usepackage{url}            
\usepackage{booktabs}       
\usepackage{amsfonts}       
\usepackage{nicefrac}       
\usepackage{microtype}      
\usepackage{xcolor}         
\usepackage{graphicx}
\usepackage{caption}
\usepackage{fancyhdr}

\title{SketchTriplet: Self-Supervised Scenarized Sketch-Text-Image Triplet Generation}

\author{
  Zhenbei Wu \\
  Beijing University of Posts and Telecommunications \\
  \texttt{wuzb@bupt.edu.cn}
  \And
  Qiang Wang\thanks{Corresponding author} \\
  Alibaba Group\\
  \texttt{yijing.wq@alibaba-inc.com} \\
  \And
  Jie Yang \\
  Beijing University of Posts and Telecommunications \\
  \texttt{janeyang@bupt.edu.cn} \\
}

\begin{document}

\maketitle

\begin{abstract}
The scarcity of free-hand sketch presents a challenging problem. Despite the emergence of some large-scale sketch datasets, these datasets primarily consist of sketches at the single-object level. There continues to be a lack of large-scale paired datasets for scene sketches. In this paper, we propose a self-supervised method for scene sketch generation that does not rely on any existing scene sketch, enabling the transformation of single-object sketches into scene sketches. To accomplish this, we introduce a method for vector sketch captioning and sketch semantic expansion. Additionally, we design a sketch generation network that incorporates a fusion of multi-modal perceptual constraints, suitable for application in zero-shot image-to-sketch downstream task, demonstrating state-of-the-art performance through experimental validation. Finally, leveraging our proposed sketch-to-sketch generation method, we contribute a large-scale dataset centered around scene sketches, comprising highly semantically consistent "text-sketch-image" triplets. Our research confirms that this dataset can significantly enhance the capabilities of existing models in sketch-based image retrieval and sketch-controlled image synthesis tasks. We will make our dataset and code publicly available.
\end{abstract}

\section{Introduction}

Free-hand sketches are a human intuitive and universal means of expressing visual information. While text represents human abstraction of semantic information, free-hand sketch represents abstraction of visual information. Compared to text, free-hand sketches are not limited by geography or culture, possessing a more universal capability for conveying information. A picture is worth a thousand words, free-hand sketches can efficiently and accurately convey visual information that is difficult to describe in language using simple lines. Free-hand sketches constitute a distinctive modality of image data generated by human hands, encapsulating human abstraction and understanding of visual information. In the field of computer vision, free-hand sketches not only enhance human efficiency in tasks such as visual design and image retrieval, but also serve as crucial tools for exploring human visual comprehension.

Currently, large-scale models have demonstrated powerful capabilities across various domains \cite{ouyang2022training, rombach2022high}, and scaling up training data remains a mainstream approach to enhancing model performance. However, the scarcity of free-hand sketch presents a challenging issue. Since free-hand sketches are exclusively created by humans, collecting sketch requires a significant amount of human effort. TU-Berlin \cite{eitz2012humans}, as the first large-scale sketch dataset, contains 20,000 sketches. Subsequently, QuickDraw \cite{ha2017neural} gamified the drawing process, resulting in the collection of 50 million sketches, making it the largest sketch dataset to date. However, both TU-Berlin and QuickDraw are single-modal sketch datasets and lack paired data from other modalities, such as natural photos and textual descriptions. Sketchy \cite{sangkloy2016sketchy} contributed a large-scale dataset with "sketch-image" pairs, comprising 75,000 paired data. However, these sketch datasets still remain at the single-object level, meaning each sketch contains only one object.

\begin{figure*}
  \centering
  \includegraphics[width=\textwidth]{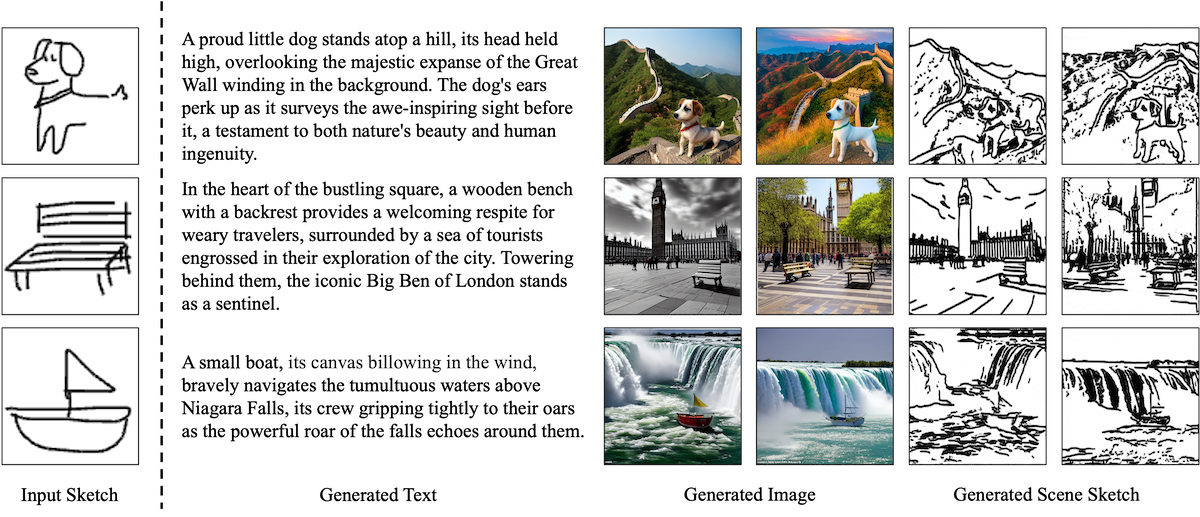}
  \caption{The scene sketch-text-image triplets produced by our method. Our approach can generate text with scene descriptions, as well as corresponding detailed images and sketches, based on the input of a single object sketch.}
  \label{fig:s2s}
\end{figure*}

With the advancement of sketch research, many researchers are shifting their focus from studying single-object-level sketches to scene-level sketches. Scene sketches typically encompass multiple objects of various categories, containing more complex abstract visual information. Scene sketches serve as vital bridges for exploring more intricate human visual comprehension and fine-grained visual design. Compared to single-object-level sketches, acquiring scene sketch is more challenging. Currently, there are three scene-level free-hand sketch datasets \cite{zou2018sketchyscene, gao2020sketchycoco, chowdhury2022fs} with paired data available. The largest in scale is SketchyCOCO \cite{gao2020sketchycoco}, containing 14,081 images. Both SketchyScene \cite{zou2018sketchyscene} and SketchyCOCO \cite{gao2020sketchycoco} employ semi-synthetic methods. However, due to the limited number of templates for single-object "sketch-image" pairs, these composite scene sketches have limitations in terms of richness. FS-COCO \cite{chowdhury2022fs} introduces the first manually drawn scene sketch dataset, consisting of 10,000 "sketch-image-text" data pairs. It is currently the most diverse and comprehensive scene sketch dataset in terms of matching modal information. However, this collection method incurs significant human and time costs.

To address these issues, we propose a self-supervised method for generating scene sketches. Our approach does not rely on any existing scene sketch but instead leverages the semantic information inherent in single-object sketches to generate scene sketches with rich elements through semantic expansion. Moreover, by adopting a multi-modal fusion approach integrating text, sketches, and images, our method can directly generalize to the task of image-to-sketch generation without the need for any "sketch-image" pairs. Leveraging this multi-modal fusion characteristic, our method simultaneously produces "text-sketch-image" triplets during the process of generating scene sketches from single-object sketches. Furthermore, these multi-modal data centered around scene sketches exhibit highly semantic consistency.

In this work, we propose three core technical strategies:
(1) Designing a GCN-based vector sketch caption method, innovatively extracting generic, foundational semantic elements from vector sketches devoid of specific attributes such as color and texture. Subsequently, the basic semantic information of the sketch is expanded into a scene sketch description by utilizing the semantic information extension method. (2) Introducing a text-driven canvas layout adaptation method to adjust the composition layout of single-object sketches, providing initialized layouts for scene sketch generation based on expanded semantic information. (3) Developing a method for scene sketch generation based on multiple constraint conditions. We propose a training approach that integrates constraints including semantic fusion perception, sketch object content perception, and multi-object perception, enabling self-supervised scene sketch generation.

\textbf{Our contributions can be summarized as follows:} 
(1) We introduce a self-supervised universal pipeline for sketch-to-sketch generation. To the best of our knowledge, it is the first method that extends single-object sketches to scene sketches. (2) We propose a method for vector sketch captioning and construct a text semantic expansion model based on sketch descriptions as the fundamental semantic elements. (3) We devise a sketch generation network based on integrating multiple modal perception constraints, enabling our proposed universal sketch-to-sketch network to easily applied to downstream zero-shot image-to-sketch tasks. Experimental results validate that it achieves state-of-the-art performance. (4) Leveraging our sketch-to-sketch network, we contribute a large-scale dataset consisting of "text-sketch-image" triplets, with scene sketches as the central component, demonstrating high semantic consistency. This dataset  fills a gap in the industry, and the dataset-driven retraining of existing models considerably enhances performance in sketch-based image retrieval and sketch-controlled image synthesis tasks.

\section{Related Work}

\textbf{Sketch Generation.}  The generation of free-hand sketches is a widely researched task. With the advancement of generative models \cite{goodfellow2014generative, sohl2015deep, ho2020denoising, song2020denoising, song2020score}, current researchers increasingly employ deep learning models to achieve sketch generation. A common approach is to generate sketches based on natural photographs, primarily using image-sketch pairs to train models for generating sketches from images. As research on free-hand sketches progresses, researchers are shifting from exploring the generation of single-object sketches to exploring the task of generating scene sketches containing multiple objects \cite{chan2022learning, yi2020unpaired, li2019photo}. However, compared to the generation from images to sketches, research on generating sketches from sketches is not sufficiently explored. Currently, the areas of focus include sketch completion \cite{su2020sketchhealer, qi2022generative, das2020beziersketch, wang2022sketchknitter} and sketch simplification \cite{mo2021general, xu2019perceptual, simo2018real, simo2018mastering, simo2016learning}. To the best of our knowledge, there has been no exploration of extending sketch generation to utilize simple single-object sketches to generate scene sketches with complex object elements.

\textbf{Diffusion Models.}  Recently, significant achievements have been made in the field of generative models based on diffusion models \cite{sohl2015deep}, \cite{ho2020denoising}, \cite{song2020denoising}, \cite{song2020score}. Thanks to the stable training objectives of diffusion models, they can better accommodate large-scale training data. Currently, diffusion models have been widely applied in various generative tasks, such as computer vision \cite{saharia2022photorealistic, rombach2022high, ramesh2022hierarchical, zhang2023adding, ruiz2023dreambooth}, natural language processing \cite{austin2021structured, li2022diffusion}, audio processing \cite{huang2022fastdiff, popov2021grad, chen2020wavegrad}, and have achieved remarkable results. Currently, customized image generation is a highly regarded research direction. Sketches, as an easy-to-use tool, are widely used to control image generation \cite{zhang2023adding, mou2023t2iadapter}. Although existing methods have successfully used simple sketches to generate impressive images, the lack of large-scale hand-drawn scene sketch datasets presents significant challenges for the task of fine-grained control in image generation using scene sketches.

\section{Methodology}

In this section, we present our method for generating scene sketches using single-object sketches. Our methodology operates independently of any existing scene sketch datasets. Rather, it leverages the semantic information inherent in single-object sketches to generate scene sketches with rich elements through semantic expansion. Furthermore, employing a multi-modal fusion technique incorporating text, sketches, and images, our method synchronously generates "text-sketch-image" triplets, during the process of generating scene sketches from single-object sketches.

\begin{figure*}
  \centering
  \includegraphics[width=\textwidth]{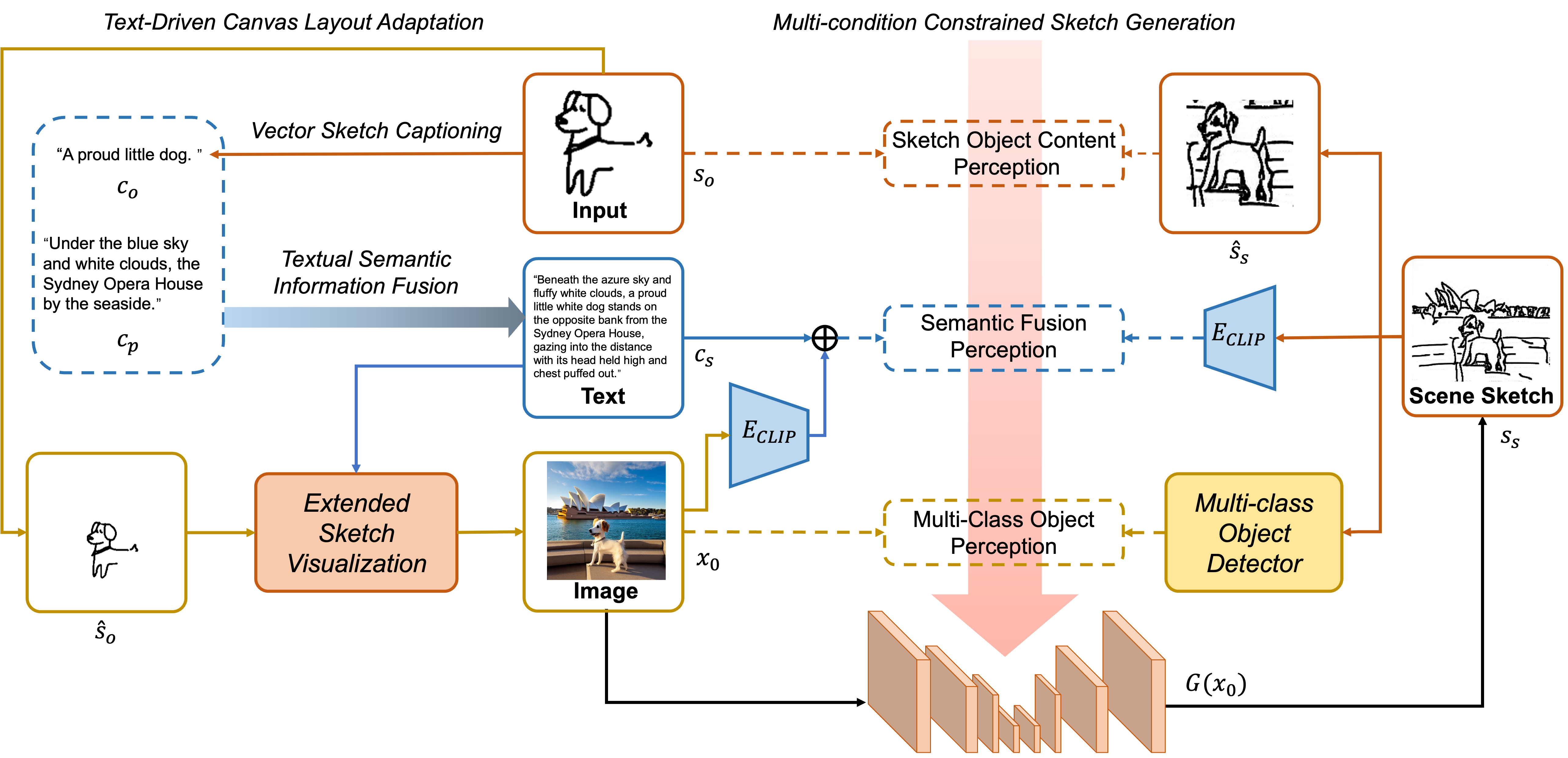}
  \caption{Pipeline overview. Through our proposed method, a skech-text-image triplet composed of scene sketch $s_s$, text $c_s$, image $x_0$ can be obtained.}
  \label{fig:overview}
\end{figure*}

Our pipeline is illustrated in Figure \ref{fig:overview}. For the input single-object sketch $s_o$, we first extract its semantic information to obtain its corresponding text representation $c_o$. Then, leveraging a large amount of prior knowledge, we perform semantic expansion on $c_o$ to obtain the extended textual semantic information form $c_s$. Finally, the single-object sketch $s_o$ and the text $c_s$ are injected into the scene sketch generation module to produce the final scene sketch $s_s$, in a self-supervised manner. Below, we will provide detailed explanations. Our method can be primarily divided into two modules: Sketch-Based Semantic Information Expansion and Scene Sketch Generation, with corresponding technical details presented in \ref{sec:sketch_exp} and \ref{sec:sketch_gen}, respectively.

\subsection{Sketch-Based Semantic Information Expansion}
\label{sec:sketch_exp}
In this section, we expand the input single-object sketch $s_o$ at the semantic level to obtain the semantically enriched textual representation $c_s$. Specifically, we employ three steps to achieve this objective. Firstly, we feed $s_o$ into a vector sketch-to-text generation model to obtain the textual representation $c_0$ of the semantic information encoded in $s_o$. Secondly, leveraging a large-scale real image dataset $P=\{p^1, p^2, \cdots, p^N\}$, where $p^i$ represents a real image and $N$ denotes the number of images in the dataset, we construct a large-scale, element-rich image description set $C_p=\{c_p^1, c_p^2, \cdots, c_p^N\}$ using an image captioning model. Here, $c_p^i$ denotes the textual description corresponding to the image $p^i$. Finally, we fuse the sketch description $c_0$ as a universal object element with the image description collection $C_p$, which contains rich background elements, to generate a scene sketch description textual collection $C_s$ with rich semantics.

\subsubsection{Vector Sketch Captioning}
\textbf{Sketch Captioning vs. Image Captioning.}  Image descriptions typically encompass color, texture, and lighting information, whereas descriptions of sketches often focus more on describing generic object shapes, owing to the inherent characteristics of sketch. In contrast to the specific referential nature of image descriptions, sketch-based descriptions tend to be more generic, representing more fundamental and universal object semantic information.

\textbf{GCN-based Vector Sketch Captioning.}  To extract sketch features from vectorized data, we construct a representation for each stroke within the graph-structured vector sketch. We define a collection of single-object sketches as $S_o=\{s_o^1, s_o^2, \cdots, s_o^M\}$. Arbitrary sketch $s_o^i$ represents a single-object sketch and $M$ denotes the number of sketches in the dataset. Each sketch using the representation proposed by Ha et al. \cite{ha2017neural}, where each point is depicted as a 3-D vector $(\Delta x, \Delta y, \Delta p)$.  Here, $(\Delta x, \Delta y)$ represents the offset distance of the pen in the $x$ and $y$ axes from the previous point, and $\Delta p$ indicates whether the stroke has ended. Building on the work of Qi et al. \cite{qi2022generative}, we developed a GCN-based encoder. We aligned the representations of the sketch and text domains using Q-Former \cite{li2023blip}. Subsequently, the output from Q-Former was integrated with Large Language Models(LLMs) to enhance the generation of visual language. Utilizing a sketch captioning model, we can generate a collection of sketch descriptions $C_o=\{c_o^1, c_o^2, \cdots, c_o^M\}$. Here, $c_o^i$ represents the description of $s_o^i$. The collection $S_o$ provides generic semantic elements for subsequent semantic expansion.

\subsubsection{Large-Scale Image Captioning}
In order to perform semantic expansion on the single-object sketch, we need to construct a large-scale collection of image-text descriptions $C_p$ as a semantic expansion repository. We have chosen a large-scale real image dataset ADE20K \cite{zhou2019semantic}, to build an element-rich image collection $P=\{p^1, p^2, \cdots, p^N\}$, where $p^i$ represents a real image from the dataset and $N$ denotes the number of images in the dataset. Subsequently, we utilize an image description model $M_c$ to generate textual descriptions $c_p^i=M_c(p^i)$ corresponding to the images $p_i$. Finally, we construct a large-scale element-rich image description collection $C_p=\{c_p^1, c_p^2, \cdots, c_p^N\}$. This image description collection $C_p$ encompasses semantic representations of a vast array of image elements, providing rich semantic elements for subsequent semantic expansion.

\subsubsection{Textual Semantic Information Fusion}
Having obtained the fundamental semantic elements from sketch-based representations and the rich semantic elements from a large corpus of images, we utilize ChatGPT \cite{ouyang2022training} to construct a fusion model for textual semantic information. With this fusion model, we integrate the basic semantic elements $c_o^i$ extracted from the single-object sketch $s_o^i$ into the rich semantic elements $c_p^j$ extracted from the image $p^j$, generating the extended textual semantic information $c_s^{(i,j)}$ for the scene sketch. Repeating this fusion process, we can build an immensely large collection of textual semantic information for scene sketches, denoted as $C_s$. By merging the sketch description collection $C_o=\{c_o^1, c_o^2, \cdots, c_o^M\}$ and the image description collection $C_p=\{c_p^1, c_p^2, \cdots, c_p^N\}$, we ultimately construct a scene sketch textual collection $C_s$ containing $M*N$ elements, represented as $C_s=\{c_s^{(1,1)}, c_s^{(1,2)}, \cdots, c_s^{(i,j)}, \cdots, c_s^{(M,N)}\}, i \in {1, 2, \cdots, M}, j \in {1, 2, \cdots, N}$. Where $\{c_s^{(i,1)}, c_s^{(i,2)}, \cdots, c_s^{(i,N)}\}$ denotes the collection of scene sketches expanded semantically based on $s_o^i$. This collection $C_s$ represents the result of semantic expansion for single-object sketches $S_o$.

\subsection{Scene Sketch Generation}
\label{sec:sketch_gen}
In this section, we utilize conditional constraints across multiple modalities to produce the final scene sketch $s_s$. In Section \ref{sec:sketch_exp}, we obtained the extended semantic information $c_s$ in textual form based on the input single-object sketch $s_o$. Leveraging the single-object sketch $s_o$ and textual semantic information $c_s$, our pipeline proceeds through three steps to obtain the final scene sketch $s_s$. (1) Using the textual semantic information $c_s$ to adjust the layout of the canvas, altering the size and position of the main object in $s_o$ to fit the object layout described in $c_s$. (2) After adjusting the canvas layout, merging the single-object sketch $s_o$ and textual semantic information $c_s$ into a large-scale generative diffusion model. Leveraging the image generation capability of  the large model, we visualize the textual semantic information $c_s$ to produce an image $x_0$. (3) We design a deep neural network model to transform the image $x_0$ into the scene sketch $s_s$. We devise a training method with multiple constraint conditions, including semantic perception, sketch object content perception, and perception of various object categories, achieving image modal transformation without the need for any paired data.

\subsubsection{Text-Driven Canvas Layout Adaptation}
\label{sec:canvas}
For a single-subject sketch, its subject matter typically occupies the center of the entire sketch and dominates most of the canvas area. In contrast, scene sketches often comprise multiple objects, with individual objects generally not occupying significant canvas space. Therefore, when expanding from a single-subject sketch $s_o$ to a scene sketch $s_s$, it is necessary not only to complete the background information of $s_o$ but also to adjust the size and position of the main subject in $s_o$. As different objects in a scene sketch exhibit certain correlations in size and position, we utilize the extended semantic information $c_s$ obtained earlier to construct a new layout for the main subject of $s_o$ within $s_s$. Following the method of Qu et al. \cite{qu2023layoutllm}, we utilize ChatGPT \cite{ouyang2022training} to generate layout coordinates that conform to the text. We define the process of scaling $s_o$ as $S_{adaptive}$, and the resulting new sketch produced after processing is denoted as $\hat{s}_o = S_{adptive}(s_0, c_s)$.

\subsubsection{Visualization of Extended Semantic Information in Sketch}
After obtaining the extended semantic information $c_s$ along with the corresponding new canvas layout, we aim to concretize the extended semantic information $c_s$ and generate visual representations of the semantic information. Leveraging the powerful image generation capability of large-scale diffusion models, we utilize $c_s$ to guide the diffusion model in generating corresponding visual images $x_0$. This visualization process helps us construct a robust "text-sketch" bridge, connecting the extended semantic information $c_s$ with the final desired scene sketch $s_s$. We employ a latent diffusion model \cite{rombach2022high}, with $s_o$ and $c_s$ as conditional guidance, to iteratively denoise $x_T \sim N(0,1)$ and reconstruct the image $x_0$. The loss function of the model is as follows, where $z_t$ represents the mapping of $x_t$ in the latent space, and $\tau_i$ and $\tau_t$ represent the encoders for the image and text respectively.

\begin{equation}
    L_{LDM}=E_{\varepsilon(x),y,\epsilon\sim N(0,1),t}[\left\|\epsilon-\epsilon_\theta(z_t,t,\tau_i(s_o)), \tau_t(c_s))\right\|_2^2]
\end{equation}

\subsubsection{Multi-condition Constrained Sketch Generation}
Due to the comprehensive integration of multiple modalities in our approach, we can generate scene sketches in a self-supervised manner, completely eliminating the dependence on "sketch-image" data pairs. Despite having only a single-object sketch $s_o$ as input, our method progressively acquires various modalities of data throughout the preceding processing stages. During the phase of semantic information expansion based on the sketch, we derive semantic expanded textual information $c_s$ from the single-object sketch $s_o$. In the visualization phase of expanded semantic information, we further obtain images $x_0$ generated by large-scale models using the text information $c_s$. Leveraging these modal information, we construct a scene sketch generation model under multiple constraint conditions to transform the image $x_0$ into the final scene sketch $s_s$. Specifically, we employ a GAN-based model, where the image $x_0$ is input into the generator $G(\cdot)$ to predict the scene sketch $s_s=G(x_0)$. Without relying on any scene sketch-image pairs, we employ three types of perception to provide constraints to the model, resulting in scene sketches that are more semantically and structurally aligned with expectations.

\textbf{Semantic Fusion Perception.}  To ensure semantic consistency among the sketch $s_s$, image $x_0$, and text $c_s$, we have employed a pre-trained CLIP model \cite{ramesh2022hierarchical}. This decision is motivated by the fact that within the CLIP model, both the image encoder and text encoder can map images and text into the same semantic space. Subsequently, it becomes feasible to gauge the semantic disparities between images and text, or between images themselves. Our objective is for the generated sketch $s_s$ to exhibit semantic alignment with the image $x_0$, while also maintaining semantic coherence between $s_s$ and the text $c_s$. Hence, we integrate these two semantic losses, defining it as $\mathcal{L}_{SFP}$, which serves to constrain the generated sketch $s_s$ to align semantically with both the text $c_s$ and image $x_0$.

\begin{equation}
    \mathcal{L}_{SFP} = ||CLIP_{image}(s_s) - CLIP_{text}(c_s)|| + ||CLIP_{image}(s_s) - CLIP_{image}(x_0)||
\end{equation}

\vspace{-0.2cm} \textbf{Sketch Object Content Perception.}  Given that our method aims to expand from a single-object sketch $s_o$ to generate a scene sketch $s_s$, there exists a strong correlation between $s_o$ and $s_s$. Despite the addition of semantic content and object elements in $s_s$ compared to $s_o$, the object of $s_o$ is retained in $s_s$. Therefore, we utilize the inverse process $\bar{S}_{adptive}(\cdot)$ of canvas expansion described in \ref{sec:canvas} to extract the portion $\hat{s}_s=\bar{S}_{adptive}(s_s, c_s)$ from $s_s$ corresponding to the main object in $s_o$. We term this disparity as $\mathcal{L}_{SOCP}$, aiming to quantify the differences between the object content in $s_o$ and $\hat{s}_s$.

\begin{equation}
    \mathcal{L}_{SOCP} = ||\bar{S}_{adptive}(s_s, c_s) - s_o||
\end{equation}

\textbf{Multi-Class Object Perception.}  Due to the complexity of multiple object compositions within scene sketches, even with overall semantic coherence, there may still exist ambiguity and confusion in the layout of objects within the images. To constrain the consistency of object layouts between $s_s$ and $x_0$, we designed a visual feature extraction backbone network based on DINOv2 \cite{oquab2023dinov2}, aimed at distinguishing different types of objects present in the images. DINOv2 possesses formidable visual perceptual capabilities, enabling effective discrimination of various types of objects within images. Specifically, we input the scene sketch into the visual feature extraction backbone network $D_m$ based on DINOv2 to obtain the visual features of the scene sketch. Subsequently, we perform principal component analysis (PCA) \cite{bro2014principal} on the extracted visual features and visualize the first three principal components. This visualization process is defined as $F_{PCA}$.

\begin{equation}
    \mathcal{L}_{MOP} = ||F_{PCA}(D_m(s_s)) - F_{PCA}(D_m(x_0))||
\end{equation}

\section{Experiments}

\subsection{Experimental Settings}
\textbf{GCN-based Vector Sketch Caption Training.}  We randomly selected $200$ classes from Quickdraw \cite{ha2017neural}, each with $1,000$ sketches, resulting in $20,000$ annotated sketch-text pairs dataset. In the first stage, we froze the LLMs and focused on training the vector sketch encoder and Q-former, ensuring that the output sketch representation could be accurately interpreted by the LLMs. During the second stage, we fine-tuned all parameters to further enhance the expressiveness of the primary objects within the sketches.

\textbf{Multi-condition constrained Sketch Generation Training.}  We randomly selected $2000$ single-object sketches from the Quickdraw \cite{ha2017neural} and 50 representative scenes from ADE20K \cite{zhou2019semantic} to construct $100,000$ sketch-image pairs using our proposed method. We employed ResNet \cite{he2016deep} blocks as the sketch generation network. For optimization, we utilized Adam to optimize with batch size $32$, a learning rate of $0.0001$, and train for $50$ epochs.

\subsection {Zero-shot Image to Sketch Generation}
\begin{figure*}
  \centering
  \includegraphics[width=\textwidth]{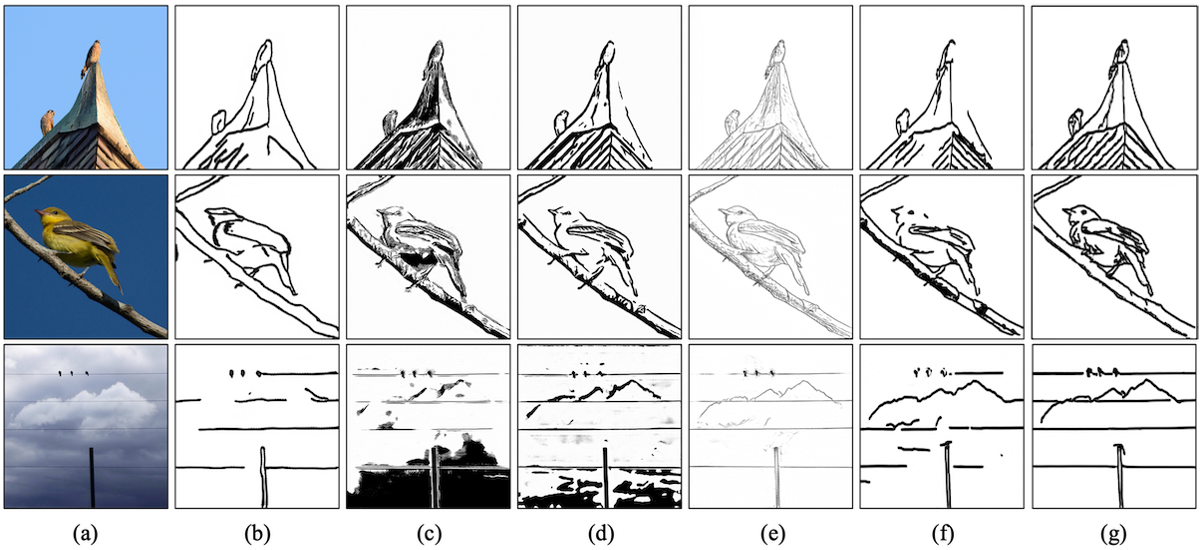}
  \caption{Comparison of results in image-to-sketch generation tasks with existing methods. (a) Input photo. (b) Results from Photo-Sketching \cite{li2019photo}. (c) Results from UPDG \cite{yi2020unpaired}, Style 1. (d) Results from UPDG \cite{yi2020unpaired}, Style 2. (e) Results from Informative-Drawings \cite{chan2022learning}, Anime style. (f) Results from Informative-Drawings \cite{chan2022learning}, Contour style. (g) Results from our method.}
  \label{fig:i2s}
\end{figure*}

Sketch is an essential interactive tool for human visual creation and editing. However, the task of generating sketches from reference images is challenging due to the scarcity of sketch, particularly fine-grained scene sketch. Fortunately, we discover that by leveraging a component of our self-supervised scene sketch generation pipeline, high-quality image-to-sketch generation can be achieved. Figure \ref{fig:i2s} illustrates the qualitative comparison results between our proposed method and other existing methods. The results showcase image-to-sketch generation for randomly selected 10,000 images from the MS-COCO \cite{lin2014microsoft}, displayed from left to right for each method. It can be observed that Photo-Sketching \cite{li2019photo} produces sketch-like lines but lacks some detail in object depiction and the accuracy of line shapes. The sketches generated by UPDG \cite{yi2020unpaired} Style 1 and Style 2 commonly exhibits issues with black shadows and includes smudging. Informative Drawings \cite{chan2022learning} style anime presents detailed sketches but suffers from excessive redundant lines. Informative-Drawings style contour offers relatively clean lines and a free-hand sketch style but still misses some critical lines. In contrast, our method produces complete and clean sketch lines. Validated by multiple evaluation metrics in Table \ref{tab:i2s}, our method achieves the best quantitative comparison results.

\begin{table*}
    \caption{Quantitative results of the scene images generated by our approach and other methods.}
    \centering
    \setlength{\tabcolsep}{3pt} 
    \begin{tabular}{ccccc}
    \hline
    & FID $\downarrow$ & IS $\uparrow$ & LPIPS $\downarrow$ & CLIP$_{iqa}$ $\uparrow$ \\
    \hline
    Photo-Sketching \cite{li2019photo} & 266.7 & 2.845 & 0.593 & 0.354 \\
    UPDG (Style1) \cite{yi2020unpaired} & 181.9 & 6.688 & 0.421 & 0.457 \\
    UPDG (Style2) \cite{yi2020unpaired} & 233.1 & 3.809 & 0.476 & 0.363 \\
    Informative-Drawings (Style Anime) \cite{chan2022learning} & 152.4 & 7.061 & 0.549 & 0.527 \\
    Informative-Drawings (Style Contour) \cite{chan2022learning} & 221.4 & 4.109 & 0.482 & 0.356 \\
    \hline
    \textit{Ours} & \bf{132.3} & \bf{8.031} & \bf{0.412} & \bf{0.542} \\
    \hline
    \end{tabular}
    \label{tab:i2s}
\end{table*}

\subsection{SketchTriplet Dataset}

\textbf{Dataset Collection.}  To the best of our knowledge, there is currently no large-scale text-sketch-image paired dataset. We have constructed a dataset, SketchTriplet, centered on sense sketch with text and image, featuring high semantic consistency. Specifically, as proposed in the pipeline shown in Figure \ref{fig:overview}, we use semantic information expansion approach introduced in \ref{sec:sketch_exp} to expand the semantic information of $s_o$ into the description $c_s$ of the scene sketch $s_s$. Next, we visualize the semantic information of the single-object sketch $s_o$ and the text $c_s$, resulting in the image $x_0$. Finally, we employed $s_o$, $c_s$, and $x_0$ as constraints to generate the scene sketch $s_s$. Through this process, we ultimately produce the triplet composed of text $c_s$, sketch $s_s$, and image $x_0$.


\textbf{Dataset Composition.}  SketchTriplet dataset consists of $1,000,000$ unique scene sketches, each accompanied by a corresponding image and a corresponding semantic label. The original categories of SketchTriplet are sourced from QuickDraw\cite{ha2017neural} with $200$ categories, each category containing $5,000$ instances. Given the inherent complexity of scene sketches, which is inevitably higher than that of single-object sketches, we reclassify our dataset into $1,000$ categories based on the image categories of ImageNet-1K \cite{deng2009imagenet}.

\textbf{Zero-shot Sketch-Based Image Retrieval.}  Sketch-based image retrieval(SBIR)\cite{chaudhuri2022bda, dutta2019semantically, shen2018zero, liu2019semantic} is a central problem in sketch understanding. Prior research efforts\cite{lin2023zero, gupta2022zeroshot, dey2019doodle, liu2019semantic} have been hampered by the formidable obstacle of assembling datasets containing aligned sketch-image-text tuples, particularly when dealing with intricate scene sketches.  In this context, we capitalized on the SketchTriplet dataset to retrain an array of advanced SBIR models, encompassing ZSE-SBIR\cite{lin2023zero}, TVT\cite{Tian2022TVTTV}, SBTKNet\cite{tursun2022efficient}, ViT-Ret\cite{dosovitskiy2020image} and SAKE\cite{liu2019semantic}.  Subsequently, these models were rigorously evaluated under a zero-shot paradigm using the TU-Berlin Extend\cite{zhang2016sketchnet}, Sketchy Extend\cite{liu2017deep} and QuickDraw Extend\cite{dey2019doodle} benchmark datasets.  As illustrated in Table \ref{tab:sbir}, each of the retrained models delivered performance that transcended the established baselines, thereby substantiating the capacity of SketchTriplet to markedly bolster the pre-existing models' competencies in both sketch interpretation and semantic analysis.

\begin{table*}
    \caption{Zero-shot scene sketch-based image retrieval results, the performance of the model after retraining is indicated in the parentheses. }
    \centering
    \setlength{\tabcolsep}{4pt} 
    \begin{tabular}{lcccccc}
    \hline
    & \multicolumn{2}{c}{TU-Berlin Extend} & \multicolumn{2}{c}{Sketchy Extend} & \multicolumn{2}{c}{QuickDraw Extend} \\
    \cmidrule(r){2-3} \cmidrule(r){4-5} \cmidrule(r){6-7}
    & mAP $\uparrow$ & Prec@100 $\uparrow$ & mAP $\uparrow$ & Prec@100 $\uparrow$ & mAP $\uparrow$ & Prec@200 $\uparrow$  \\
    \hline ZSE-SBIR& 0.60(1.7\%) & 0.64(2.7\%) & 0.74(2.0\%) & 0.81(1.6\%) & 0.14(14.7\%) & 0.20(4.2\%)\\
    TVT & 0.49(3.3\%) & 0.66(3.9\%) & 0.65(5.3\%) & 0.80(1.8\%) & 0.15(15.0\%) & 0.29(9.2\%) \\
    SBTKNet & 0.48(4.4\%) & 0.61(4.6\%) & 0.55(6.1\%) & 0.70(2.9\%) & 0.13(14.2\%) & 0.34(2.5\%)  \\
    SAKE & 0.48(8.2\%) & 0.60(4.4\%) & 0.55(6.3\%) & 0.70(3.5\%) & 0.13(11.4\%) & 0.18(3.4\%) \\
    ViT-Ret & 0.44(5.4\%) & 0.58(4.1\%) & 0.48(6.5\%) & 0.64(3.1\%) & 0.12(10.0\%) & 0.13(8.4\%)\\
    \hline
    \end{tabular}
    \label{tab:sbir}
\end{table*}

\begin{figure*}
  \centering
  \includegraphics[width=\textwidth]{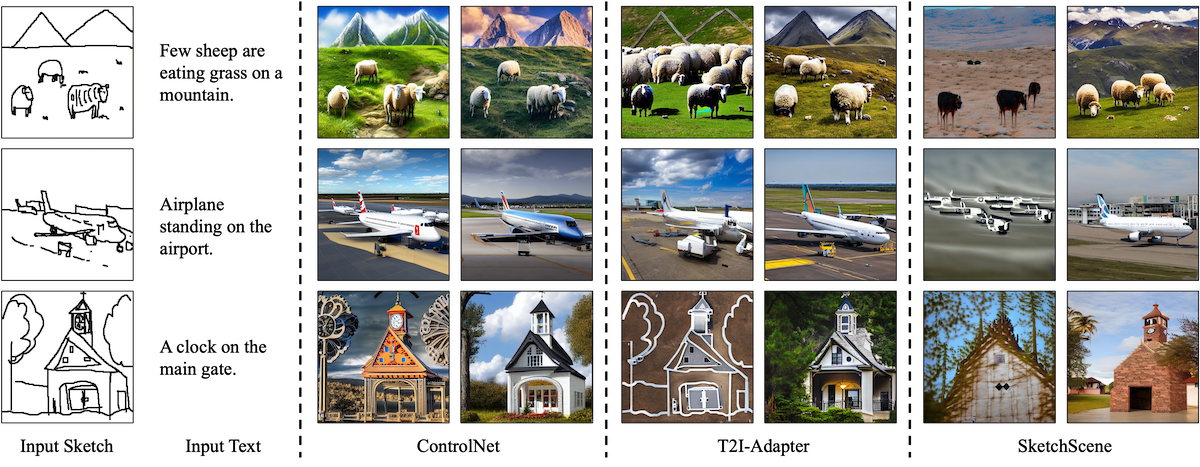}
  \caption{Qualitative results of existing scene sketch-controlled image generation methods on scene sketch dataset FS-COCO\cite{chowdhury2022fs}. Within the same method, the images in the left and right columns represent the results before and after further training with our SketchTriplet dataset.}
  \label{fig:sketch_to_image}
\end{figure*}

\textbf{Sketch Control Image Synthesis.}  Controlling image synthesis through sketches\cite{wang2023diffsketching, wu2023sketchscene, isola2017image} has always been a classic theme in cross-modal conversion, requiring reliance on large-scale sketch-to-image paired datasets. This task grows particularly formidable when attempting to recreate images from sketches depicting intricate scenes, where the objective is to reconstruct images with a high degree of detail. We conducted experiments employing the advanced methods, including ControlNet\cite{zhang2023adding}, T2I-Adapter\cite{mou2023t2iadapter} and SketchScene\cite{wu2023sketchscene}. The outcomes of our experiments are meticulously documented in Figure \ref{fig:sketch_to_image} and Table \ref{tab:sketch_to_image}. Generation models based on Diffusion Models are typical data-driven approaches\cite{dhariwal2021diffusion}, and SketchTriplet has greatly enhanced the capabilities of existing methods.

\begin{table*}
    \caption{Quantitative results of scene sketch to image generation, the performance of the model after retraining is indicated in the parentheses.}
    \centering
    \setlength{\tabcolsep}{7pt} 
    \begin{tabular}{ccccc}
    \hline
    & FID $\downarrow$ & IS $\uparrow$ & LPIPS $\downarrow$ & CLIP$_{iqa}$ $\uparrow$ \\
    \hline
    ControlNet \cite{zhang2023adding} & 38.76(7.8\%) & 10.05(6.9\%) & 0.492(6.1\%) & 0.835(2.5\%) \\
    T2I-Adapter \cite{mou2023t2iadapter} & 45.06(8.4\%) & 9.187(9.0\%) & 0.481(4.2\%) & 0.854(1.4\%) \\
    SketchScene \cite{wu2023sketchscene} & 49.09(5.1\%) & 8.803(9.1\%) & 0.541(5.6\%) & 0.717(4.9\%) \\
    \hline
    \end{tabular}
    \label{tab:sketch_to_image}
\end{table*}

\subsection{Ablation Study}
\begin{figure*}
  \centering
  \includegraphics[width=\textwidth]{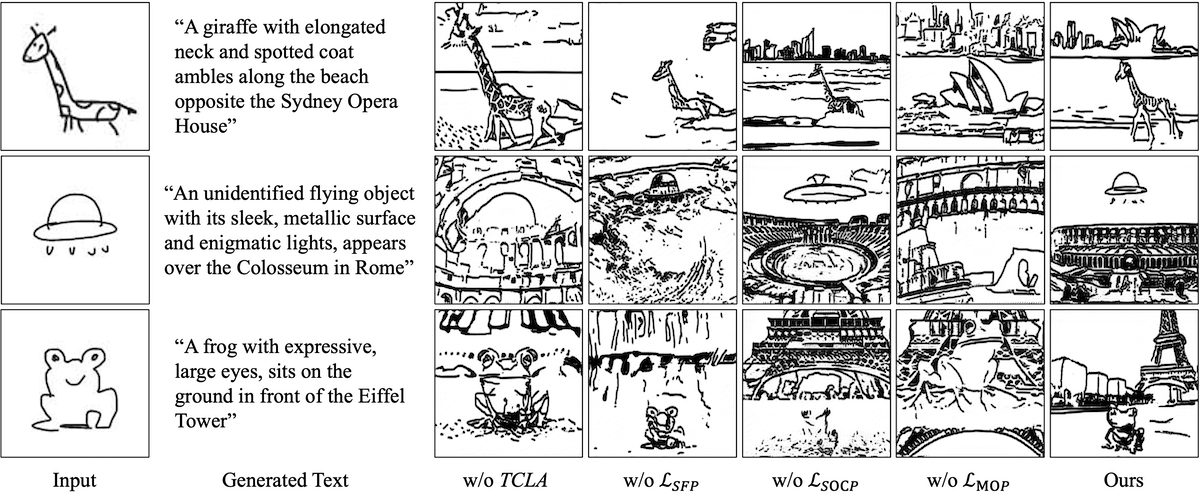}
  \caption{Results of the ablation study. The first and second columns on the left show the input single-object sketch and the text description generated by our model. The third to sixth columns represent the results of removing the Text-driven Canvas Layout Adaptation module $TCLA$ and the three constraint losses.}
  \label{fig:ablation}
\end{figure*}

Figure \ref{fig:ablation} shows the results of the ablation study for our method. Firstly, we removed the Text-driven Canvas Layout Adaptation module $TCLA$. It can be observed that while the generated scene sketches contain core objects and outlines of some background elements, the background elements are not recognizable because the main spatial area is occupied. Next, we conducted ablation studies on the three losses used to constrain sketch generation. When the semantic fusion perception loss $L_{SFP}$ was removed, the resulting scene sketches only contain outlines similar to the input sketch, with disordered lines and largely unrecognizable objects. Upon removing the $L_{SOCP}$, based on sketch object content perception, the generated scene sketches align with the description but exhibit missing and blurred lines in the main parts of the input sketch. After eliminating the $L_{MOP}$, which accounts for multi-object perception, the generated scene sketches include appropriate object elements, but their spatial distribution is chaotic.

\section{Conclusion}
In this paper, we propose a self-supervised scene sketch generation method. Our approach can extend single-object sketches into scene sketches without relying on any scene sketch. We innovatively introduce a vector sketch captioning method and design a sketch generation network based on multi-modal perception constraints. Using our proposed method, we can efficiently produce scenarized "sketch-text-image" triplet, ultimately contributing a large-scale scene sketch dataset. This significantly addresses the current scarcity of paired free-hand scene sketch, and our method is poised to become a data-driven force for sketch-based applications.

\textbf{Limitation.}  Our approach has two limitations. First, our sketch generation method does not provide control over transparency. A possible solution might be to design brush transparency as a learnable parameter. Second, our method produces sketches in only one fixed style. A potential solution could be to enhance the model's capability to learn different sketching styles.

\small
\bibliographystyle{plainnat}
\bibliography{arxiv}

\begin{thebibliography}{59}
\providecommand{\natexlab}[1]{#1}
\providecommand{\url}[1]{\texttt{#1}}
\expandafter\ifx\csname urlstyle\endcsname\relax
  \providecommand{\doi}[1]{doi: #1}\else
  \providecommand{\doi}{doi: \begingroup \urlstyle{rm}\Url}\fi

\bibitem[Austin et~al.(2021)Austin, Johnson, Ho, Tarlow, and Van Den~Berg]{austin2021structured}
Jacob Austin, Daniel~D Johnson, Jonathan Ho, Daniel Tarlow, and Rianne Van Den~Berg.
\newblock Structured denoising diffusion models in discrete state-spaces.
\newblock \emph{Advances in Neural Information Processing Systems}, 34:\penalty0 17981--17993, 2021.

\bibitem[Bro and Smilde(2014)]{bro2014principal}
Rasmus Bro and Age~K Smilde.
\newblock Principal component analysis.
\newblock \emph{Analytical methods}, 6\penalty0 (9):\penalty0 2812--2831, 2014.

\bibitem[Chan et~al.(2022)Chan, Durand, and Isola]{chan2022learning}
Caroline Chan, Fr{\'e}do Durand, and Phillip Isola.
\newblock Learning to generate line drawings that convey geometry and semantics.
\newblock In \emph{Proceedings of the IEEE/CVF Conference on Computer Vision and Pattern Recognition}, pages 7915--7925, 2022.

\bibitem[Chaudhuri et~al.(2022)Chaudhuri, Chavan, Banerjee, Dutta, and Akata]{chaudhuri2022bda}
Ushasi Chaudhuri, Ruchika Chavan, Biplab Banerjee, Anjan Dutta, and Zeynep Akata.
\newblock Bda-sketret: Bi-level domain adaptation for zero-shot sbir.
\newblock \emph{Neurocomputing}, 514:\penalty0 245--255, 2022.

\bibitem[Chen et~al.(2020)Chen, Zhang, Zen, Weiss, Norouzi, and Chan]{chen2020wavegrad}
Nanxin Chen, Yu~Zhang, Heiga Zen, Ron~J Weiss, Mohammad Norouzi, and William Chan.
\newblock Wavegrad: Estimating gradients for waveform generation.
\newblock \emph{arXiv preprint arXiv:2009.00713}, 2020.

\bibitem[Chowdhury et~al.(2022)Chowdhury, Sain, Bhunia, Xiang, Gryaditskaya, and Song]{chowdhury2022fs}
Pinaki~Nath Chowdhury, Aneeshan Sain, Ayan~Kumar Bhunia, Tao Xiang, Yulia Gryaditskaya, and Yi-Zhe Song.
\newblock Fs-coco: Towards understanding of freehand sketches of common objects in context.
\newblock In \emph{European Conference on Computer Vision}, pages 253--270. Springer, 2022.

\bibitem[Das et~al.(2020)Das, Yang, Hospedales, Xiang, and Song]{das2020beziersketch}
Ayan Das, Yongxin Yang, Timothy Hospedales, Tao Xiang, and Yi-Zhe Song.
\newblock B{\'e}ziersketch: A generative model for scalable vector sketches.
\newblock In \emph{Computer Vision--ECCV 2020: 16th European Conference, Glasgow, UK, August 23--28, 2020, Proceedings, Part XXVI 16}, pages 632--647. Springer, 2020.

\bibitem[Deng et~al.(2009)Deng, Dong, Socher, Li, Li, and Fei-Fei]{deng2009imagenet}
Jia Deng, Wei Dong, Richard Socher, Li-Jia Li, Kai Li, and Li~Fei-Fei.
\newblock Imagenet: A large-scale hierarchical image database.
\newblock In \emph{2009 IEEE conference on computer vision and pattern recognition}, pages 248--255. Ieee, 2009.

\bibitem[Dey et~al.(2019)Dey, Riba, Dutta, Llados, and Song]{dey2019doodle}
Sounak Dey, Pau Riba, Anjan Dutta, Josep Llados, and Yi-Zhe Song.
\newblock Doodle to search: Practical zero-shot sketch-based image retrieval.
\newblock In \emph{Proceedings of the IEEE/CVF conference on computer vision and pattern recognition}, pages 2179--2188, 2019.

\bibitem[Dhariwal and Nichol(2021)]{dhariwal2021diffusion}
Prafulla Dhariwal and Alexander Nichol.
\newblock Diffusion models beat gans on image synthesis.
\newblock \emph{Advances in neural information processing systems}, 34:\penalty0 8780--8794, 2021.

\bibitem[Dosovitskiy et~al.(2020)Dosovitskiy, Beyer, Kolesnikov, Weissenborn, Zhai, Unterthiner, Dehghani, Minderer, Heigold, Gelly, et~al.]{dosovitskiy2020image}
Alexey Dosovitskiy, Lucas Beyer, Alexander Kolesnikov, Dirk Weissenborn, Xiaohua Zhai, Thomas Unterthiner, Mostafa Dehghani, Matthias Minderer, Georg Heigold, Sylvain Gelly, et~al.
\newblock An image is worth 16x16 words: Transformers for image recognition at scale.
\newblock \emph{arXiv preprint arXiv:2010.11929}, 2020.

\bibitem[Dutta and Akata(2019)]{dutta2019semantically}
Anjan Dutta and Zeynep Akata.
\newblock Semantically tied paired cycle consistency for zero-shot sketch-based image retrieval.
\newblock In \emph{Proceedings of the IEEE/CVF Conference on Computer Vision and Pattern Recognition}, pages 5089--5098, 2019.

\bibitem[Eitz et~al.(2012)Eitz, Hays, and Alexa]{eitz2012humans}
Mathias Eitz, James Hays, and Marc Alexa.
\newblock How do humans sketch objects?
\newblock \emph{ACM Transactions on graphics (TOG)}, 31\penalty0 (4):\penalty0 1--10, 2012.

\bibitem[Gao et~al.(2020)Gao, Liu, Xu, Wang, Liu, and Zou]{gao2020sketchycoco}
Chengying Gao, Qi~Liu, Qi~Xu, Limin Wang, Jianzhuang Liu, and Changqing Zou.
\newblock Sketchycoco: Image generation from freehand scene sketches.
\newblock In \emph{Proceedings of the IEEE/CVF conference on computer vision and pattern recognition}, pages 5174--5183, 2020.

\bibitem[Goodfellow et~al.(2014)Goodfellow, Pouget-Abadie, Mirza, Xu, Warde-Farley, Ozair, Courville, and Bengio]{goodfellow2014generative}
Ian~J. Goodfellow, Jean Pouget-Abadie, Mehdi Mirza, Bing Xu, David Warde-Farley, Sherjil Ozair, Aaron Courville, and Yoshua Bengio.
\newblock Generative adversarial networks, 2014.

\bibitem[Gupta et~al.(2022)Gupta, Chaudhuri, and Banerjee]{gupta2022zeroshot}
Sumrit Gupta, Ushasi Chaudhuri, and Biplab Banerjee.
\newblock Zero-shot sketch based image retrieval using graph transformer, 2022.

\bibitem[Ha and Eck(2017)]{ha2017neural}
David Ha and Douglas Eck.
\newblock A neural representation of sketch drawings.
\newblock \emph{arXiv preprint arXiv:1704.03477}, 2017.

\bibitem[He et~al.(2016)He, Zhang, Ren, and Sun]{he2016deep}
Kaiming He, Xiangyu Zhang, Shaoqing Ren, and Jian Sun.
\newblock Deep residual learning for image recognition.
\newblock In \emph{Proceedings of the IEEE conference on computer vision and pattern recognition}, pages 770--778, 2016.

\bibitem[Ho et~al.(2020)Ho, Jain, and Abbeel]{ho2020denoising}
Jonathan Ho, Ajay Jain, and Pieter Abbeel.
\newblock Denoising diffusion probabilistic models.
\newblock In \emph{Advances in Neural Information Processing Systems}, volume~33, pages 6840--6851, 2020.

\bibitem[Huang et~al.(2022)Huang, Lam, Wang, Su, Yu, Ren, and Zhao]{huang2022fastdiff}
Rongjie Huang, Max~WY Lam, Jun Wang, Dan Su, Dong Yu, Yi~Ren, and Zhou Zhao.
\newblock Fastdiff: A fast conditional diffusion model for high-quality speech synthesis.
\newblock \emph{arXiv preprint arXiv:2204.09934}, 2022.

\bibitem[Isola et~al.(2017)Isola, Zhu, Zhou, and Efros]{isola2017image}
Phillip Isola, Jun-Yan Zhu, Tinghui Zhou, and Alexei~A Efros.
\newblock Image-to-image translation with conditional adversarial networks.
\newblock In \emph{Proceedings of the IEEE conference on computer vision and pattern recognition}, pages 1125--1134, 2017.

\bibitem[Li et~al.(2023)Li, Li, Savarese, and Hoi]{li2023blip}
Junnan Li, Dongxu Li, Silvio Savarese, and Steven Hoi.
\newblock Blip-2: Bootstrapping language-image pre-training with frozen image encoders and large language models.
\newblock In \emph{International conference on machine learning}, pages 19730--19742. PMLR, 2023.

\bibitem[Li et~al.(2019)Li, Lin, Mech, Yumer, and Ramanan]{li2019photo}
Mengtian Li, Zhe Lin, Radomir Mech, Ersin Yumer, and Deva Ramanan.
\newblock Photo-sketching: Inferring contour drawings from images.
\newblock In \emph{2019 IEEE Winter Conference on Applications of Computer Vision (WACV)}, pages 1403--1412. IEEE, 2019.

\bibitem[Li et~al.(2022)Li, Thickstun, Gulrajani, Liang, and Hashimoto]{li2022diffusion}
Xiang Li, John Thickstun, Ishaan Gulrajani, Percy~S Liang, and Tatsunori~B Hashimoto.
\newblock Diffusion-lm improves controllable text generation.
\newblock \emph{Advances in Neural Information Processing Systems}, 35:\penalty0 4328--4343, 2022.

\bibitem[Lin et~al.(2023)Lin, Li, Li, Hospedales, Song, and Qi]{lin2023zero}
Fengyin Lin, Mingkang Li, Da~Li, Timothy Hospedales, Yi-Zhe Song, and Yonggang Qi.
\newblock Zero-shot everything sketch-based image retrieval, and in explainable style.
\newblock In \emph{Proceedings of the IEEE/CVF Conference on Computer Vision and Pattern Recognition}, pages 23349--23358, 2023.

\bibitem[Lin et~al.(2014)Lin, Maire, Belongie, Hays, Perona, Ramanan, Doll{\'a}r, and Zitnick]{lin2014microsoft}
Tsung-Yi Lin, Michael Maire, Serge Belongie, James Hays, Pietro Perona, Deva Ramanan, Piotr Doll{\'a}r, and C~Lawrence Zitnick.
\newblock Microsoft coco: Common objects in context.
\newblock In \emph{Computer Vision--ECCV 2014: 13th European Conference, Zurich, Switzerland, September 6-12, 2014, Proceedings, Part V 13}, pages 740--755. Springer, 2014.

\bibitem[Liu et~al.(2017)Liu, Shen, Shen, Liu, and Shao]{liu2017deep}
Li~Liu, Fumin Shen, Yuming Shen, Xianglong Liu, and Ling Shao.
\newblock Deep sketch hashing: Fast free-hand sketch-based image retrieval.
\newblock In \emph{Proceedings of the IEEE conference on computer vision and pattern recognition}, pages 2862--2871, 2017.

\bibitem[Liu et~al.(2019)Liu, Xie, Wang, and Yuille]{liu2019semantic}
Qing Liu, Lingxi Xie, Huiyu Wang, and Alan~L Yuille.
\newblock Semantic-aware knowledge preservation for zero-shot sketch-based image retrieval.
\newblock In \emph{Proceedings of the IEEE/CVF International Conference on Computer Vision}, pages 3662--3671, 2019.

\bibitem[Mo et~al.(2021)Mo, Simo-Serra, Gao, Zou, and Wang]{mo2021general}
Haoran Mo, Edgar Simo-Serra, Chengying Gao, Changqing Zou, and Ruomei Wang.
\newblock General virtual sketching framework for vector line art.
\newblock \emph{ACM Transactions on Graphics (TOG)}, 40\penalty0 (4):\penalty0 1--14, 2021.

\bibitem[Mou et~al.(2023)Mou, Wang, Xie, Wu, Zhang, Qi, Shan, and Qie]{mou2023t2iadapter}
Chong Mou, Xintao Wang, Liangbin Xie, Yanze Wu, Jian Zhang, Zhongang Qi, Ying Shan, and Xiaohu Qie.
\newblock T2i-adapter: Learning adapters to dig out more controllable ability for text-to-image diffusion models, 2023.

\bibitem[Oquab et~al.(2023)Oquab, Darcet, Moutakanni, Vo, Szafraniec, Khalidov, Fernandez, Haziza, Massa, El-Nouby, et~al.]{oquab2023dinov2}
Maxime Oquab, Timoth{\'e}e Darcet, Th{\'e}o Moutakanni, Huy Vo, Marc Szafraniec, Vasil Khalidov, Pierre Fernandez, Daniel Haziza, Francisco Massa, Alaaeldin El-Nouby, et~al.
\newblock Dinov2: Learning robust visual features without supervision.
\newblock \emph{arXiv preprint arXiv:2304.07193}, 2023.

\bibitem[Ouyang et~al.(2022)Ouyang, Wu, Jiang, Almeida, Wainwright, Mishkin, Zhang, Agarwal, Slama, Ray, et~al.]{ouyang2022training}
Long Ouyang, Jeffrey Wu, Xu~Jiang, Diogo Almeida, Carroll Wainwright, Pamela Mishkin, Chong Zhang, Sandhini Agarwal, Katarina Slama, Alex Ray, et~al.
\newblock Training language models to follow instructions with human feedback.
\newblock \emph{Advances in neural information processing systems}, 35:\penalty0 27730--27744, 2022.

\bibitem[Popov et~al.(2021)Popov, Vovk, Gogoryan, Sadekova, and Kudinov]{popov2021grad}
Vadim Popov, Ivan Vovk, Vladimir Gogoryan, Tasnima Sadekova, and Mikhail Kudinov.
\newblock Grad-tts: A diffusion probabilistic model for text-to-speech.
\newblock In \emph{International Conference on Machine Learning}, pages 8599--8608. PMLR, 2021.

\bibitem[Qi et~al.(2022)Qi, Su, Wang, Yang, Pang, and Song]{qi2022generative}
Yonggang Qi, Guoyao Su, Qiang Wang, Jie Yang, Kaiyue Pang, and Yi-Zhe Song.
\newblock Generative sketch healing.
\newblock \emph{International Journal of Computer Vision}, 130\penalty0 (8):\penalty0 2006--2021, 2022.

\bibitem[Qu et~al.(2023)Qu, Wu, Fei, Nie, and Chua]{qu2023layoutllm}
Leigang Qu, Shengqiong Wu, Hao Fei, Liqiang Nie, and Tat-Seng Chua.
\newblock Layoutllm-t2i: Eliciting layout guidance from llm for text-to-image generation.
\newblock In \emph{Proceedings of the 31st ACM International Conference on Multimedia}, pages 643--654, 2023.

\bibitem[Ramesh et~al.(2022)Ramesh, Dhariwal, Nichol, Chu, and Chen]{ramesh2022hierarchical}
Aditya Ramesh, Prafulla Dhariwal, Alex Nichol, Casey Chu, and Mark Chen.
\newblock Hierarchical text-conditional image generation with clip latents.
\newblock \emph{arXiv preprint arXiv:2204.06125}, 1\penalty0 (2):\penalty0 3, 2022.

\bibitem[Rombach et~al.(2022)Rombach, Blattmann, Lorenz, Esser, and Ommer]{rombach2022high}
Robin Rombach, Andreas Blattmann, Dominik Lorenz, Patrick Esser, and Bj{\"o}rn Ommer.
\newblock High-resolution image synthesis with latent diffusion models.
\newblock In \emph{Proceedings of the IEEE/CVF conference on computer vision and pattern recognition}, pages 10684--10695, 2022.

\bibitem[Ruiz et~al.(2023)Ruiz, Li, Jampani, Pritch, Rubinstein, and Aberman]{ruiz2023dreambooth}
Nataniel Ruiz, Yuanzhen Li, Varun Jampani, Yael Pritch, Michael Rubinstein, and Kfir Aberman.
\newblock Dreambooth: Fine tuning text-to-image diffusion models for subject-driven generation.
\newblock In \emph{Proceedings of the IEEE/CVF Conference on Computer Vision and Pattern Recognition}, pages 22500--22510, 2023.

\bibitem[Saharia et~al.(2022)Saharia, Chan, Saxena, Li, Whang, Denton, Ghasemipour, Gontijo~Lopes, Karagol~Ayan, Salimans, et~al.]{saharia2022photorealistic}
Chitwan Saharia, William Chan, Saurabh Saxena, Lala Li, Jay Whang, Emily~L Denton, Kamyar Ghasemipour, Raphael Gontijo~Lopes, Burcu Karagol~Ayan, Tim Salimans, et~al.
\newblock Photorealistic text-to-image diffusion models with deep language understanding.
\newblock \emph{Advances in Neural Information Processing Systems}, 35:\penalty0 36479--36494, 2022.

\bibitem[Sangkloy et~al.(2016)Sangkloy, Burnell, Ham, and Hays]{sangkloy2016sketchy}
Patsorn Sangkloy, Nathan Burnell, Cusuh Ham, and James Hays.
\newblock The sketchy database: learning to retrieve badly drawn bunnies.
\newblock \emph{ACM Transactions on Graphics (TOG)}, 35\penalty0 (4):\penalty0 1--12, 2016.

\bibitem[Shen et~al.(2018)Shen, Liu, Shen, and Shao]{shen2018zero}
Yuming Shen, Li~Liu, Fumin Shen, and Ling Shao.
\newblock Zero-shot sketch-image hashing.
\newblock In \emph{Proceedings of the IEEE conference on computer vision and pattern recognition}, pages 3598--3607, 2018.

\bibitem[Simo-Serra et~al.(2016)Simo-Serra, Iizuka, Sasaki, and Ishikawa]{simo2016learning}
Edgar Simo-Serra, Satoshi Iizuka, Kazuma Sasaki, and Hiroshi Ishikawa.
\newblock Learning to simplify: fully convolutional networks for rough sketch cleanup.
\newblock \emph{ACM Transactions on Graphics (TOG)}, 35\penalty0 (4):\penalty0 1--11, 2016.

\bibitem[Simo-Serra et~al.(2018{\natexlab{a}})Simo-Serra, Iizuka, and Ishikawa]{simo2018mastering}
Edgar Simo-Serra, Satoshi Iizuka, and Hiroshi Ishikawa.
\newblock Mastering sketching: adversarial augmentation for structured prediction.
\newblock \emph{ACM Transactions on Graphics (TOG)}, 37\penalty0 (1):\penalty0 1--13, 2018{\natexlab{a}}.

\bibitem[Simo-Serra et~al.(2018{\natexlab{b}})Simo-Serra, Iizuka, and Ishikawa]{simo2018real}
Edgar Simo-Serra, Satoshi Iizuka, and Hiroshi Ishikawa.
\newblock Real-time data-driven interactive rough sketch inking.
\newblock \emph{ACM Transactions on Graphics (TOG)}, 37\penalty0 (4):\penalty0 1--14, 2018{\natexlab{b}}.

\bibitem[Sohl-Dickstein et~al.(2015)Sohl-Dickstein, Weiss, Maheswaranathan, and Ganguli]{sohl2015deep}
Jascha Sohl-Dickstein, Eric Weiss, Niru Maheswaranathan, and Surya Ganguli.
\newblock Deep unsupervised learning using nonequilibrium thermodynamics.
\newblock In \emph{International Conference on Machine Learning}, pages 2256--2265. PMLR, 2015.

\bibitem[Song et~al.(2020{\natexlab{a}})Song, Meng, and Ermon]{song2020denoising}
Jiaming Song, Chenlin Meng, and Stefano Ermon.
\newblock Denoising diffusion implicit models.
\newblock In \emph{International Conference on Learning Representations}, 2020{\natexlab{a}}.

\bibitem[Song et~al.(2020{\natexlab{b}})Song, Sohl-Dickstein, Kingma, Kumar, Ermon, and Poole]{song2020score}
Yang Song, Jascha Sohl-Dickstein, Diederik~P Kingma, Abhishek Kumar, Stefano Ermon, and Ben Poole.
\newblock Score-based generative modeling through stochastic differential equations.
\newblock \emph{arXiv preprint arXiv:2011.13456}, 2020{\natexlab{b}}.

\bibitem[Su et~al.(2020)Su, Qi, Pang, Yang, and Song]{su2020sketchhealer}
Guoyao Su, Yonggang Qi, Kaiyue Pang, Jie Yang, and Yi-Zhe Song.
\newblock Sketchhealer a graph-to-sequence network for recreating partial human sketches.
\newblock In \emph{Proceedings of The 31st British Machine Vision Virtual Conference (BMVC 2020)}, pages 1--14. British Machine Vision Association, 2020.

\bibitem[Tian et~al.(2022)Tian, Xu, Shen, Yang, and Shen]{Tian2022TVTTV}
Jialin Tian, Xing Xu, Fumin Shen, Yang Yang, and Heng~Tao Shen.
\newblock Tvt: Three-way vision transformer through multi-modal hypersphere learning for zero-shot sketch-based image retrieval.
\newblock In \emph{AAAI Conference on Artificial Intelligence}, 2022.
\newblock URL \url{https://api.semanticscholar.org/CorpusID:249536611}.

\bibitem[Tursun et~al.(2022)Tursun, Denman, Sridharan, Goan, and Fookes]{tursun2022efficient}
Osman Tursun, Simon Denman, Sridha Sridharan, Ethan Goan, and Clinton Fookes.
\newblock An efficient framework for zero-shot sketch-based image retrieval.
\newblock \emph{Pattern Recognition}, 126:\penalty0 108528, 2022.

\bibitem[Wang et~al.(2022)Wang, Deng, Qi, Li, and Song]{wang2022sketchknitter}
Qiang Wang, Haoge Deng, Yonggang Qi, Da~Li, and Yi-Zhe Song.
\newblock Sketchknitter: Vectorized sketch generation with diffusion models.
\newblock In \emph{The Eleventh International Conference on Learning Representations}, 2022.

\bibitem[Wang et~al.(2023)Wang, Kong, Lin, and Qi]{wang2023diffsketching}
Qiang Wang, Di~Kong, Fengyin Lin, and Yonggang Qi.
\newblock Diffsketching: Sketch control image synthesis with diffusion models.
\newblock \emph{arXiv preprint arXiv:2305.18812}, 2023.

\bibitem[Wu et~al.(2023)Wu, Deng, Wang, Kong, Yang, and Qi]{wu2023sketchscene}
Zhenbei Wu, Haoge Deng, Qiang Wang, Di~Kong, Jie Yang, and Yonggang Qi.
\newblock Sketchscene: Scene sketch to image generation with diffusion models.
\newblock In \emph{2023 IEEE International Conference on Multimedia and Expo (ICME)}, pages 2087--2092. IEEE, 2023.

\bibitem[Xu et~al.(2019)Xu, Xie, Miao, Qu, Xiao, Zhang, Liu, and Wong]{xu2019perceptual}
Xuemiao Xu, Minshan Xie, Peiqi Miao, Wei Qu, Wenpeng Xiao, Huaidong Zhang, Xueting Liu, and Tien-Tsin Wong.
\newblock Perceptual-aware sketch simplification based on integrated vgg layers.
\newblock \emph{IEEE transactions on visualization and computer graphics}, 27\penalty0 (1):\penalty0 178--189, 2019.

\bibitem[Yi et~al.(2020)Yi, Liu, Lai, and Rosin]{yi2020unpaired}
Ran Yi, Yong-Jin Liu, Yu-Kun Lai, and Paul~L Rosin.
\newblock Unpaired portrait drawing generation via asymmetric cycle mapping.
\newblock In \emph{Proceedings of the IEEE/CVF conference on computer vision and pattern recognition}, pages 8217--8225, 2020.

\bibitem[Zhang et~al.(2016)Zhang, Liu, Zhang, Ren, Wang, and Cao]{zhang2016sketchnet}
Hua Zhang, Si~Liu, Changqing Zhang, Wenqi Ren, Rui Wang, and Xiaochun Cao.
\newblock Sketchnet: Sketch classification with web images.
\newblock In \emph{Proceedings of the IEEE conference on computer vision and pattern recognition}, pages 1105--1113, 2016.

\bibitem[Zhang et~al.(2023)Zhang, Rao, and Agrawala]{zhang2023adding}
Lvmin Zhang, Anyi Rao, and Maneesh Agrawala.
\newblock Adding conditional control to text-to-image diffusion models.
\newblock In \emph{Proceedings of the IEEE/CVF International Conference on Computer Vision}, pages 3836--3847, 2023.

\bibitem[Zhou et~al.(2019)Zhou, Zhao, Puig, Xiao, Fidler, Barriuso, and Torralba]{zhou2019semantic}
Bolei Zhou, Hang Zhao, Xavier Puig, Tete Xiao, Sanja Fidler, Adela Barriuso, and Antonio Torralba.
\newblock Semantic understanding of scenes through the ade20k dataset.
\newblock \emph{International Journal of Computer Vision}, 127:\penalty0 302--321, 2019.

\bibitem[Zou et~al.(2018)Zou, Yu, Du, Mo, Song, Xiang, Gao, Chen, and Zhang]{zou2018sketchyscene}
Changqing Zou, Qian Yu, Ruofei Du, Haoran Mo, Yi-Zhe Song, Tao Xiang, Chengying Gao, Baoquan Chen, and Hao Zhang.
\newblock Sketchyscene: Richly-annotated scene sketches.
\newblock In \emph{Proceedings of the european conference on computer vision (ECCV)}, pages 421--436, 2018.

\end{thebibliography}

\end{document}